\documentclass[10pt,twocolumn,letterpaper]{article}

\usepackage{cvpr}
\usepackage{times}
\usepackage{epsfig}
\usepackage{graphicx}
\usepackage{amsmath}
\usepackage{amssymb}
\usepackage{multirow}
\usepackage{booktabs} 
\usepackage{mathrsfs}
\usepackage{bm}
\usepackage{resizegather}
\usepackage{pifont}
\usepackage{authblk}
\newcommand{\cmark}{{\color{blue}\ding{51}}}%
\newcommand{\xmark}{{\color{red}\ding{55}}}%

\usepackage[pagebackref=true,breaklinks=true,letterpaper=true,colorlinks,bookmarks=false]{hyperref}
\graphicspath{{figures/}}
\cvprfinalcopy 

\def\cvprPaperID{****} 

\DeclareFontFamily{U}{mathc}{}
\DeclareFontShape{U}{mathc}{m}{it}%
{<->s*[1.03] mathc10}{}

\DeclareMathAlphabet{\mathscr}{U}{mathc}{m}{it}

\def\sexyname{REVERIE\xspace}
\newcommand{\vect}[1]{\mathbf{#1}}
\newcommand{\overrightarrowa}[1]{\vect{#1}}  
\newcommand{\tuple}[1]{\mathscr{#1}}

\ifcvprfinal\fi
\begin{document}
	
	\title{REVERIE: Remote Embodied Visual Referring Expression in \\Real Indoor Environments}
	
	\author[1]{Yuankai Qi}
	\author[1]{Qi Wu\thanks{corresponding author; qi.wu01@adelaide.edu.au}}
	\author[2]{Peter Anderson}
	\author[3]{Xin Wang}
	\author[3]{William Yang Wang}
	\author[1]{Chunhua Shen}
	\author[1]{Anton van den Hengel}
	\affil[1]{Australia Centre for Robotic Vision, The University of Adelaide}
	\affil[2]{Georgia Institute of Technology}
	\affil[3]{University of California, Santa Barbara}

	\maketitle

	\begin{abstract}
		
		One of the long-term challenges of robotics is to enable robots to interact with humans in the visual world via natural language, as humans are visual animals that communicate through language.
		Overcoming this challenge  requires the ability to perform a wide variety of complex tasks in response to multifarious instructions from humans.
		In the hope that it might drive progress towards more flexible and powerful human interactions with robots, we propose a dataset of varied and complex robot tasks, described in natural language, in terms of objects visible in a large set of real images.
		Given an instruction, success requires navigating through a previously-unseen environment to identify an object. This represents a practical challenge, but one that closely reflects one of the core visual problems in robotics.
		Several state-of-the-art vision-and-language navigation, and referring-expression models are tested to verify the difficulty of this new task, but none of them show promising results because there are many fundamental differences between our task and previous ones.
		A novel Interactive Navigator-Pointer model is also proposed that provides a strong baseline on the task. The proposed model especially achieves the best performance on the unseen test split, but still leaves substantial room for improvement compared to the human performance.
	\end{abstract}
	
	\section{Introduction} 

\begin{figure}[!t]
	\begin{center}
		\includegraphics[width=\linewidth]{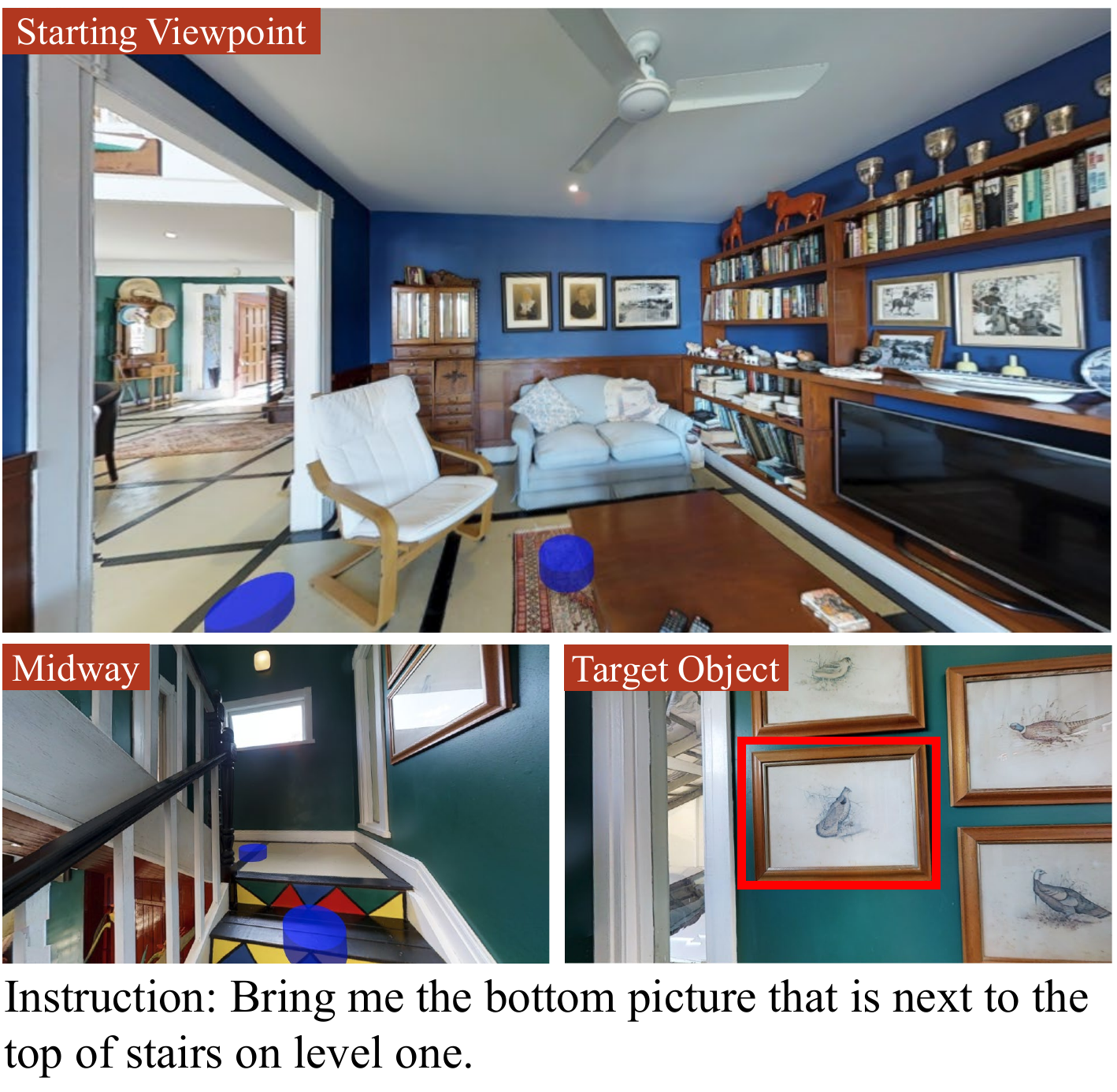}
	\end{center}
	\vspace{-4mm}
	\caption{\sexyname task: an agent is given a natural language instruction referring to a remote object  (here in the red bounding box) in a photo-realistic 3D environment. The agent must navigate to an appropriate location and identify the object from multiple distracting candidates. 
		The blue discs indicate nearby navigable viewpoints provided by the simulator.
	}
	\label{fig:title}
\end{figure}

You can ask a 10-year-old child to bring you a cushion, and there is a good chance that they will succeed (even in an unfamiliar environment), while the probability that a robot will achieve the same task is significantly lower.  
Children have a wealth of knowledge learned from similar environments that they can easily apply to such tasks in an unfamiliar environment, including the facts that cushions generally inhabit couches, that couches inhabit lounge rooms, and that lounge rooms are often connected to the rest of a building through hallways.
Children are also able to interpret natural language instructions and associate them with the visual world.  
However, the fact that robots currently lack these capabilities dramatically limits their domain of application.

\begin{table*}[!t]
\centering
\resizebox{\linewidth}{!}{
	\begin{tabular}{l}
		\toprule
		1. Fold the towel in the bathroom with the fishing theme. \\
		2. Enter the bedroom with the letter E over the bed and turn the light switch off. \\
		3. Go to the blue family room and bring the framed picture of a person on a horse at the top left  corner above the TV. \\
		4. Push in the bar chair, in the kitchen, by the oven. \\
		5. Windex the mirror above the sink, in the bedroom with the large, stone fireplace. \\
		6. Could you please dust the light above the toilet in the bathroom that is near the entry way? \\
		7. At the top of the stairs, the first set of potted flowers in front of the stairs need to be dusted off. \\
		8. To the right at the end of the hall, where the large blue table foot stool is, there is a mirror that needs to be wiped. \\
		9. Go to the hallway area where there are three pictures side by side and get me the one on the right. \\
		10. There is a bottle in the office alcove next to the piano. It is on the shelf above the sink on the extreme right. Please bring it here.  \\
		\bottomrule
\end{tabular}}%
\vspace{2mm}
\caption{Indicative instruction examples from the \sexyname dataset illustrating various interesting linguistic phenomena such as dangling modifiers (\eg 1),  spatial relations (\eg 3), imperatives (\eg 9),  co-references (\eg 10), \etc. Note that the agent in our task is required to identify the referent object, but is not required to complete any manipulation tasks (such as folding the towel).}
\vspace{-3mm}
\label{tab:challengeInstr}%
\end{table*}%

Therefore, to equip robots with such abilities and to advance real-world vision-and-language research, we introduce a new problem, which we refer to as \emph{Remote Embodied Visual referring Expression in Real Indoor Environments} --- \emph{\sexyname}. 
An example of the \sexyname task is illustrated in Fig.~\ref{fig:title}.  
A robot spawns at a starting location and is given a natural language instruction that refers to a remote target object at another location within the same building.
To carry out the task, the agent is required to navigate closer to the object and return a bounding box encompassing the target object specified by the instruction.
It demands the robot to infer the probable location of the object using knowledge of the environments, and explicitly identify the object according to the language instruction.

In distinction to other embodied tasks such as Vision-and-Language Navigation (VLN)~\cite{r2r} and Embodied Question Answering (EQA)~\cite{eqa}, REVEVIE evaluates the success based on explicit object grounding rather than the point navigation in VLN or the question answering in EQA. 
This more clearly reflects the necessity of robots' capability of natural language understanding, visual navigation, and object grounding.   
More importantly, the concise instructions in REVERIE represent more practical tasks that humans would ask a robot to perform (see Tab.~\ref{tab:challengeInstr}). Those high-level instructions are fundamentally different from the fine-grained visuomotor instructions in VLN, and would empower high-level reasoning and real-world applications. 
Moreover, compared to the task of Referring Expression (RefExp) \cite{referitgame,mao2016generation} that selects the desired object from a single image, REVERIE is far more challenging in the sense that the target object is not visible in the initial view and needs to be discovered by actively navigating in the environment. Hence, in REVERIE, there are at least an order of magnitude more object candidates to choose from than RefExp.

We build the REVERIE dataset upon the Matterport3D Simulator~\cite{r2r,mp3d}, which provides panoramas of all the navigable locations and the connectivity graph in a building. 
To provide object-level information of the environments, we have extended the simulator to incorporate object annotations, including labels and bounding boxes from Chang~\etal~\cite{mp3d}. 
The extended simulator can project bounding boxes onto images of different viewpoints and angles, thus able to accommodate evaluation on every possible location.   
The \sexyname dataset comprises 10,567 panoramas within 90 buildings containing 4,140 target objects, and 21,702 crowd-sourced instructions with an average length of 18 words. Tab.~\ref{tab:challengeInstr} demonstrates sample instructions from the dataset, which illustrate various linguistic phenomena, including spatial relations, multiple long and dangling modifiers, and coreferences, etc.

We investigate the difficulty of the \sexyname{} task by directly combining state-of-the-art (SoTA) navigation methods and referring expression methods, and none of them shows promising results.
We then propose an Interactive Navigator-Pointer model serving as a strong baseline for the \sexyname{} task.
We also provide the human performance of the REVERIE task on the test set to quantify the machine-human gap.

In summary, our main contributions are:
\begin{enumerate}
\item A new embodied vision-and-language problem, Remote Embodied Visual referring Expressions in Real 3D Indoor Environments (REVERIE), where given a natural language instruction that represents  a practical task to perform, an agent must navigate and identify a remote object in real indoor environments. 
\item 
The first benchmark dataset for the REVERIE task, which contains large-scale human-annotated instructions and extends the Matterport3D Simulator \cite{r2r} with additional object annotations.
\item A novel interactive navigator-pointer model that provides strong baselines for the \sexyname{} dataset under several evaluation metrics. 
\end{enumerate}
	
	\begin{table*}[!ht]
  \centering
  \resizebox{0.9\linewidth}{!}{
    \begin{tabular}{l|cccc|ccc|c}
    \toprule
    \multirow{2}[0]{*}{\textbf{Dataset}}      & \multicolumn{4}{c|}{\textbf{Language Context}} & \multicolumn{3}{c|}{\textbf{Visual Context}} & \multirow{2}[0]{*}{\textbf{Goal}} \\
          & Human & Main Content & Unamb   & Guidance Level & BBox & Real-world & Temporal &  \\
    \hline
    EQA \cite{eqa}, IQA \cite{iqa} & \xmark     & QA-pair & \cmark     & --    & \xmark     & \xmark     & Dynamic & QA \\
    MARCO \cite{marco}, DRIF \cite{drif} & \cmark     & Nav-Instruction   & \cmark     & Detailed & \xmark     & \xmark     & Dynamic & Navigation \\
    R2R \cite{r2r}   & \cmark     & Nav-Instruction   & \cmark     & Detailed & \xmark     & \cmark     & Dynamic & Navigation \\
    TouchDown \cite{chen2019touchdown} & \cmark     & Nav-Instruction   & \cmark     & Detailed & \xmark     & \cmark     & Dynamic & Navigation \\
    VLNA \cite{nguyen2019vision}, HANNA\cite{nguyen2019help} & \xmark     & Nav-Dialog & \xmark     & High   & \xmark     & \cmark     & Dynamic & Find Object \\
    TtW \cite{ttw}   & \cmark     & Nav-Dialog & \cmark     & High  & \xmark     & \cmark     & Dynamic & Navigation \\
    CVDN \cite{vdn} &\cmark & Nav-Dialog& \xmark & High & \xmark & \cmark & Dynamic & Find Room \\
    ReferCOCO \cite{YuPYBB16} & \cmark     & RefExp & \cmark     & --    & \cmark     & \cmark     & Static & Localise Object \\
    \hline
    REVERIE & \cmark     & Remote RefExp & \cmark     & High  & \cmark     & \cmark     & Dynamic & Localise Remote Object \\
    \bottomrule
    \end{tabular}}%
\vspace{2mm}
\caption{Compared to existing datasets involving embodied vision and language tasks. Symbol instruction: `QA': `Question-Answer', `Unamb': `Unambiguous', `BBox': `Bounding Box', `Dynamic'/`Static': visual context temporally changed or not. }
\vspace{-2mm}
\label{tab:datacmp}%
\end{table*}%

\vspace{-3pt}
\section{Related Work}

\paragraph{Referring Expression Comprehension.}
The referring expression comprehension task requires an agent to localise an object in an image given a natural language expression. 
Recent work casts this task as looking for the object that can generate its paired expressions~\cite{HuXRFSD16,Luos17,YuPYBB16}  or jointly embedding the image and expression for matching estimation~\cite{ChenKN17,HuRADS17,Liu0017,mattnet}.                         
Yu \etal~\cite{YuPYBB16} propose to compute the appearance difference of the same category objects to enhance the visual features for expression generation. 
Instead of treating each expression as a unit,~\cite{mattnet} learns to decompose an expression into appearance, location, and object relationship three components.  

Different from referring expression, \sexyname{} introduces three new challenges: i) The refereed object is not visible in the initial scene and only can be accessed after navigating to a closed location. 
ii) In contrast to previous referring expression tasks that select the target object from a single image, object candidates in \sexyname{} come from panoramas of all the possible viewpoints.
iii) The objects in referring expression are normally captured from the front view, while in our setting, the visual appearances of objects may vary largely due to different observation angles and viewpoints.

\vspace{-10pt}
\paragraph{Vision-and-Language Navigation.}
Vision-and-language navigation (VLN) is the task where an agent is to navigate to a goal location in a 3D simulator given detailed natural language instructions such as ``Turn right and go through the kitchen. Walk past the couches on the right and into the hallway on the left. Go straight until you get to a room that is to the left of the pictures of children on the wall. Turn left and go into the bathroom. Wait near the sink."~\cite{r2r}. 
A range of VLN methods~\cite{vlnNips,fast,regret,selfMonitor,cross,vlnEccv} have been proposed to address this VLN task. 

Although the proposed \sexyname{} task also requires an agent to navigate to a goal location, it differs from existing VLN tasks in two important aspects: i) {The challenge is much more closely related to the overarching objective of enabling natural language robot tasking } because the goal is to localise a target object specified in an instruction, not just a location.
This removes the artificial constraint that the instruction is restricted to solely to navigation, and reflects the reality of the fact that most objects can be seen from multiple viewpoints. 
ii) Our collected navigation instructions are semantic-level commands which better reflect the way humans communicate.  
They are thus closer to `the cold tap in the first bedroom on level two' rather than step by step navigation instructions such as `go to the top of the stairs then turn left and walk along the hallway and stop at the first bedroom on your right'.

The most closely related challenge to that proposed here is that addressed in \cite{nguyen2019help,nguyen2019vision,vdn} whereby an agent must identify an object by requesting  and interpreting  natural language assistance.  The instructions are of the form `Find a mug', and the assumption is that there is an oracle following the agent around the environment willing to provide natural language assistance. The question is then whether the agent can effectively exploit the assistance provided by the omniscient oracle. \sexyname{}, in contrast, evaluates whether the agent can carry out a natural-language instruction alone. 
Another closely related work is TOUCHDOWN \cite{chen2019touchdown}, that requires an agent to find a location in an urban outdoor environment on the basis of detailed navigation instructions.

\vspace{-8pt}
\paragraph{Embodied Question Answering.}
Embodied question answering (EQA)~\cite{eqa} requires an agent to answer a question about an object or a room in a synthetic environment. Gordon \etal~\cite{iqa} introduce an interactive version of the EQA task, where the agent may need to interact with the environment/objects to correctly answer questions. 
Our \sexyname{} task differs from previous works that only output a simple answer or a series of actions, as we ask the agent to output a bounding box around a target object. This is a more challenging but realistic setting because if we want a robot to carry out a task that relates to an object, we need its precise location. Tab.~\ref{tab:datacmp} displays the difference between our task and other related embodied vision-language tasks.
	
\section{The \sexyname{} Dataset}
We now describe the \sexyname{} task and dataset, including the task definition, evaluation metrics, simulator, data collection policy, and analysis of the collected instructions. 

\subsection{The \sexyname{} Task}
As shown in Fig. \ref{fig:title}, our \sexyname{} task requires an intelligent agent to correctly localise a remote target object (can not be observed at starting location) specified by a concise high-level natural language instruction (see samples in Tab.~\ref{tab:challengeInstr}). 
It is worth noting that our instructions are much closer to practical scenarios in daily life than the detailed instructions in VLN~\cite{r2r}, because the latter ones are so complex and long that are unrealistic for human to command robots.
Since the target object is in a different room from the starting one, the agent needs first to navigate to the goal location.

{Formally, at the beginning of each episode, the agent is given as input a high-level natural language instruction $\mathcal{X}=\langle w_1,w_2,\cdots,w_L\rangle$, where $L$ is the length of the instruction and  $w_i$ is a single word token. Following the common practice in VLN, the agent has access to surrounding panoramic images $\mathcal{V}_0=\{v_{0,k}, k\in1,\dots,36\}$ and navigable viewpoints at the current location, where $v_{0,k}$ is determined by the agent's states comprising a tuple of 3D position, heading and elevation $\tuple{s}_{0,k}=\langle p_0,\phi_{0,k},\theta_{0,k}\rangle$ (3 elevation and 12 heading angles are used). Then the agent needs to make a sequence of actions $\langle a_0,\cdots,a_T\rangle$ to reach the goal location, where each action is choosing one of the navigable viewpoints or choosing the current viewpoint which means to stop. 
The action can also be a `detecting' action that outputs the target object bounding-box refereed by the instruction.
It is worth noting that the agent can attempt to localise the target at any step, which is totally up to algorithm design. But we only allow the agent output once in each episode, which means the agent only can guess the answer once in a single run.
If the agent `thinks' it has localised the target object and decides to output it, it is required to output  a bounding box or choose from several candidates provided by the simulator. 
A bounding box is denoted as $\langle b_x,b_y,b_w,b_h\rangle$, where $b_x$ and  $b_y$ are the coordinate of the left-top point, $b_w$ and $b_h$ denote the width and height of the bounding box, respectively. 
The episode ends after the agent outputs the target bounding box.}

\subsection{Evaluation Metrics}
\label{sec:evmetrics}
The performance of a model is mainly measured by \sexyname{} success rate, which is the number of successful tasks over the total number of tasks. A task is considered successful if it selects the correct bounding box of the target object from a set of candidates
(or the IoU between the predicted bounding box and the ground-truth bounding box $\geq 0.5$, when candidate objects bounding boxes are not given). Because the target object can be observed at different viewpoints or camera views, we treat it as a success as long as the agent can identify the target within 3 meters, regardless of from different viewpoints or views.
We also measure the navigation performance with four kinds of metrics, including success rate, oracle success rate, success rate weighted by path length (SPL), and path length (in meters)~\cite{r2r}. Please note that in our task, a navigation is considered successful only when the agent stops at a location within 3 meters from the target object. More details of evaluation metrics can be found in supplementary materials.

\subsection{The \sexyname{} Simulator}
Our simulator is based on the Matterport3D Simulator~\cite{r2r}, a large-scale interactive environment constructed from the Matterport3D dataset~\cite{mp3d}.
In the simulator, an embodied agent is able to virtually `move' throughout each building by iteratively selecting adjacent nodes from the graph of panoramic viewpoints and adjusting the camera pose at each viewpoint. At each viewpoint, it returns a rendered colour image that captures the current view, as shown in Fig.~\ref{fig:title}.

\label{sec:addbbox}
\begin{figure}[t]
\begin{center}
\includegraphics[width=1\linewidth]{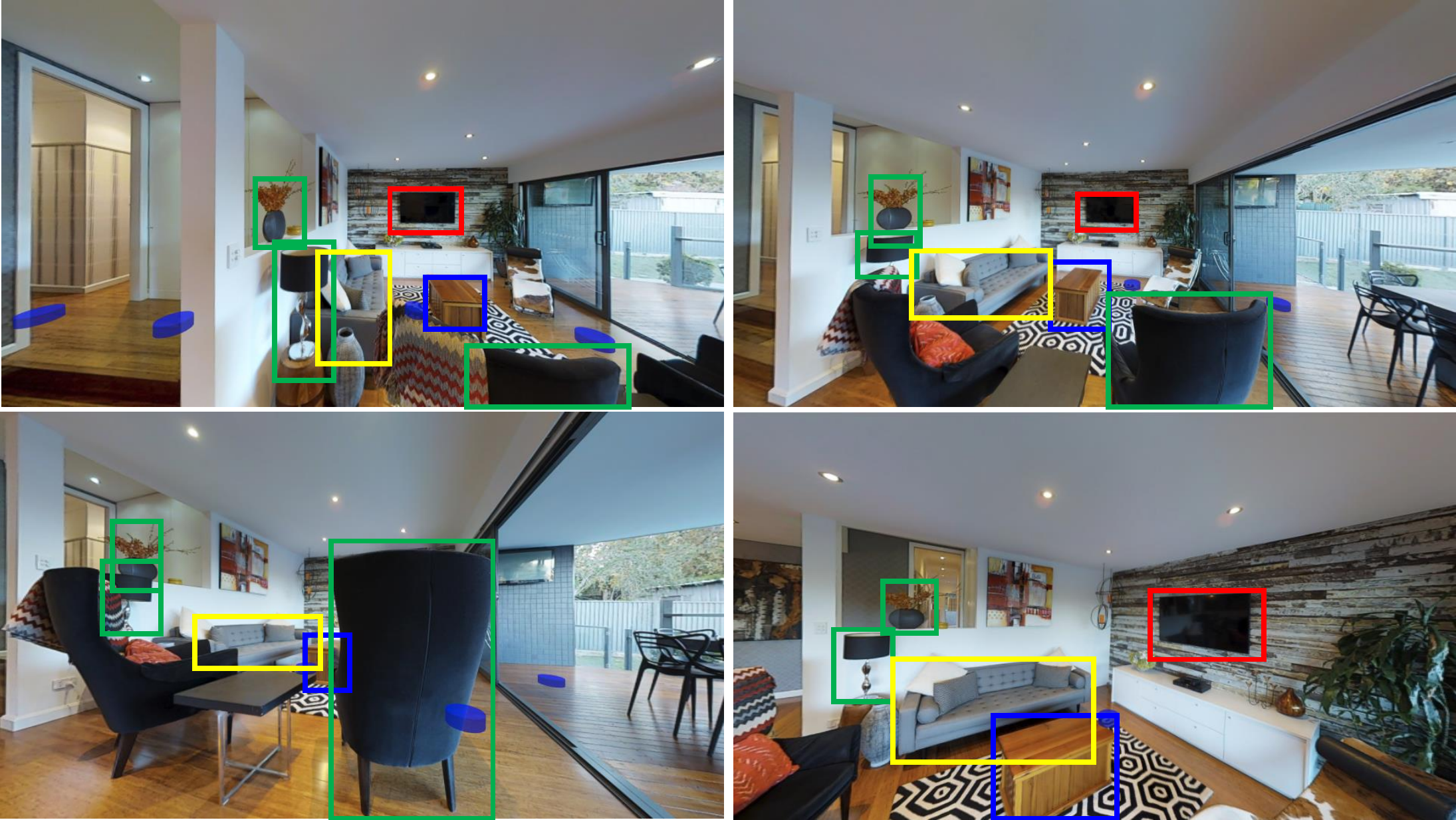}
\end{center}
\vspace{-5pt}
\caption{Object bounding boxes (BBox) in our simulator. The BBox size and aspect ratio of the same object may change after the agent moves to another viewpoint or changes its camera view. }
\label{fig:cmp2R2R}
\vspace{-12pt}
\end{figure}

\vspace{-10pt}
\paragraph{Adding Object-level Annotations.}
Object bounding boxes are needed in our proposed task, which are either provided as object hypotheses or used to assess the agent's ability to localise the object that is referred to by a natural expression. 
The main challenge of adding the object bounding boxes into the simulator is that we need to handle the changes in visibility and coordinate of 2D bounding boxes as the camera moves or rotates.

To address these issues, we calculate the overlap between bounding boxes and object depth in each view. If a bounding box is fully covered by another one and it has a larger depth, we treat it as an occluded case. Specifically, for each building the Matterport3D dataset provides all the objects appearing in it with centre point position $\overrightarrowa{c}=\langle c_x,c_y,c_z\rangle$, three axis directions 
$\overrightarrowa{d}_i=\langle d_i^x,d_i^y,d_i^z\rangle, 
i\in \{1,2,3\}$, and three radii $r_i$, one for each axis direction.
To correctly render objects in the web simulator, we first calculate the eight vertexes using $\overrightarrowa{c},\overrightarrowa{d}_i$ and $r_i$. Then these vertexes are projected into the camera space by the camera pose provided by Matterport3D dataset. %
Both C++ and web simulators will be released with the code.
Fig. \ref{fig:cmp2R2R} presents an example of projected bounding boxes. %
Note that the target object may be observed at multiple viewpoints in one room, but we expect a robot can reach the target in a short distance. Thus, we only preserve objects within three meters to a viewpoint. For each object,  {a class label and a bounding box are} associated. But it is worth noting that we adjust the size and aspect-ratio accordingly as the viewpoint and camera angle change.
In total, we obtain 20k object annotations.

\subsection{Data Collection}
Our goal is to collect high-level human daily commands that may be assigned to a home robot in future, such as  `Open the left window in the kitchen' or `Go to my bedroom and bring me a pillow'. 
We develop an interactive 3D WebGL simulator to collect such instructions on Amazon Mechanical Turk (AMT). The web simulator  first shows a path animation and then randomly highlights one object at the goal location for workers to provide instructions to find or operate with. 
There is no style limitation of the command as long as it can lead the robot to the target object. Assistant room and object information are provided to workers facilitating them to provide unambiguous instructions if there are similar rooms or objects.
The workers can look around at the goal location to learn the environment. 
For each target object, we collect three referring expressions. 
The full collection interface (see in supplementary) is the result of several rounds of experimentation. Over {1,000 workers} took part in the data collection, totally contributing around {2,648 hours} of annotation time. Examples of the collected data can be found in Tab.~\ref{tab:challengeInstr}, and more are in supplementary.

\begin{figure}[t]
    \centering
    \resizebox{1.04\linewidth}{!}{
    \begin{tabular}{cc}
        \hspace{-10pt} \includegraphics[width=0.5\linewidth]{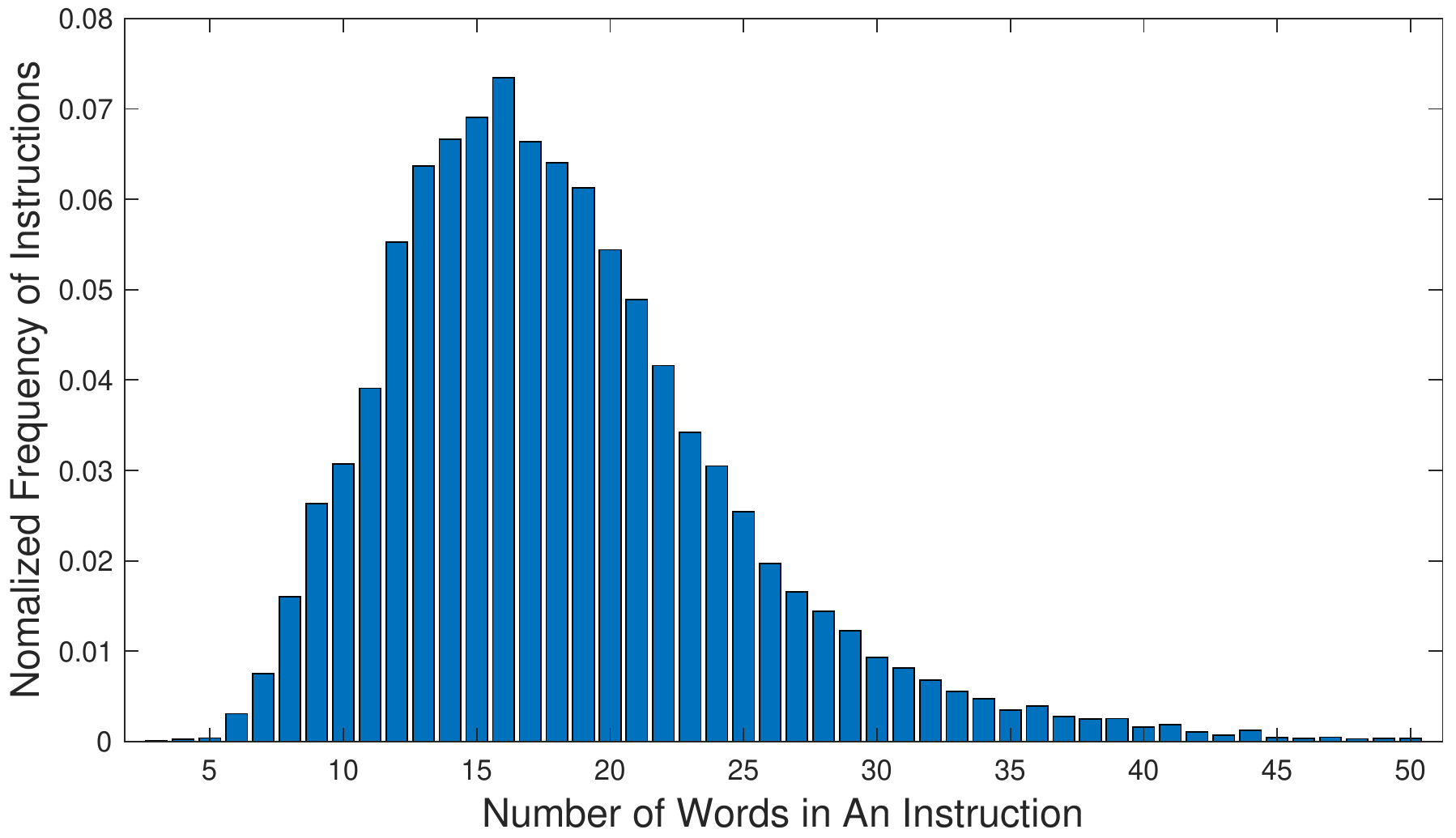}&
         \hspace{-10pt}\includegraphics[width=0.5\linewidth]{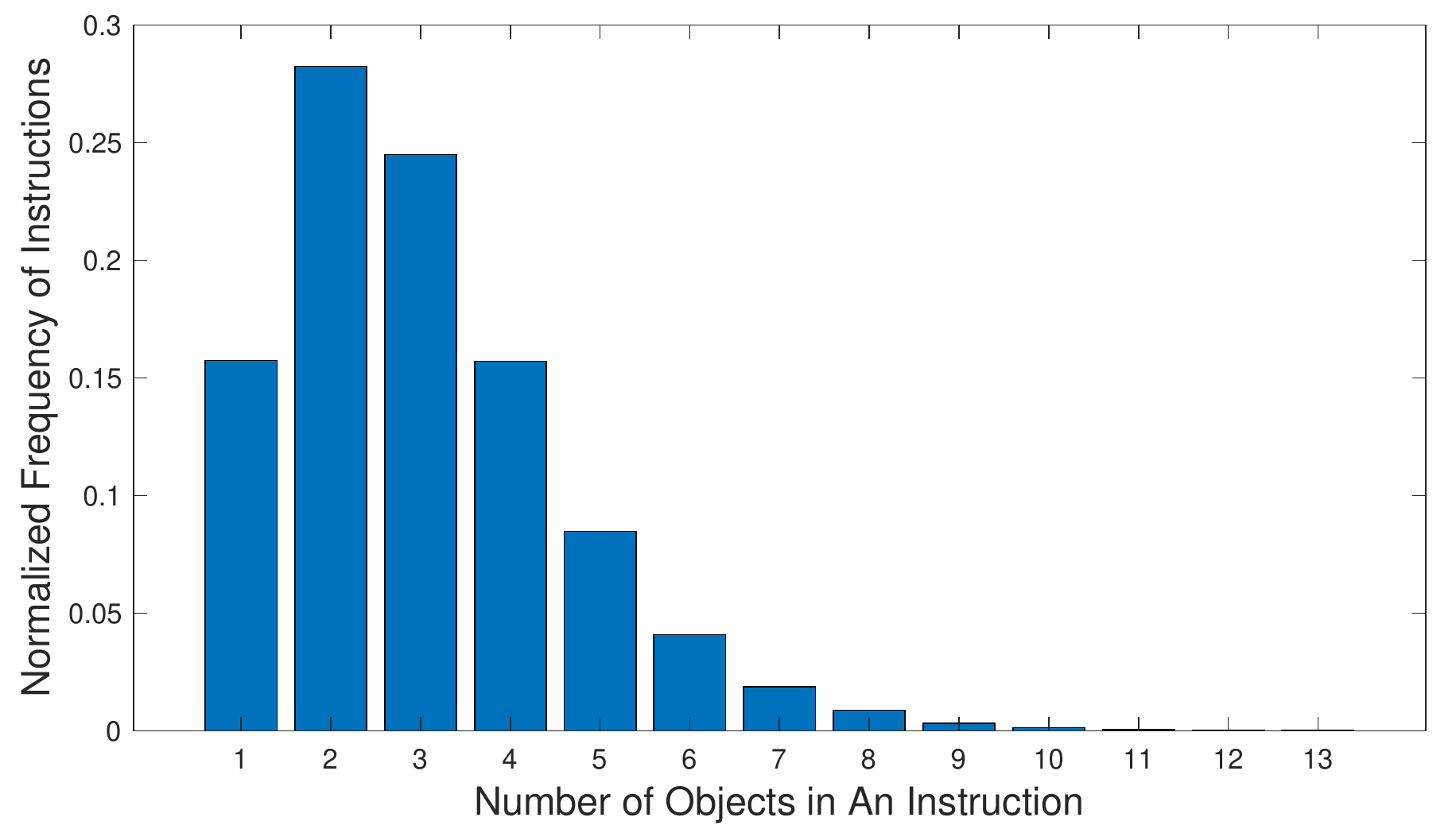}
    \end{tabular}}
\caption{The distribution of the number of words (left) and objects (right) in each instruction.}
\vspace{-2mm}
\label{fig:distri}
\end{figure}

\subsection{Dataset Analysis}

The \sexyname{} dataset contains totally 21,702 instructions and a vocabulary of over 1,600 words. The average length of collected instructions is 18 words involving both navigation and referring expression information. 
Considering the detailed navigation instructions provided in the R2R~\cite{r2r} with an average length of 29 words and the previous largest dataset RefCOCOg \cite{YuPYBB16} contains an average of 8 words, our instructions command is much more concise and natural, and thus more challenging.

\begin{figure}[t]
    \centering
    \resizebox{1.04\linewidth}{!}{
    \begin{tabular}{cc}
        \hspace{-10pt} \includegraphics[width=0.5\linewidth]{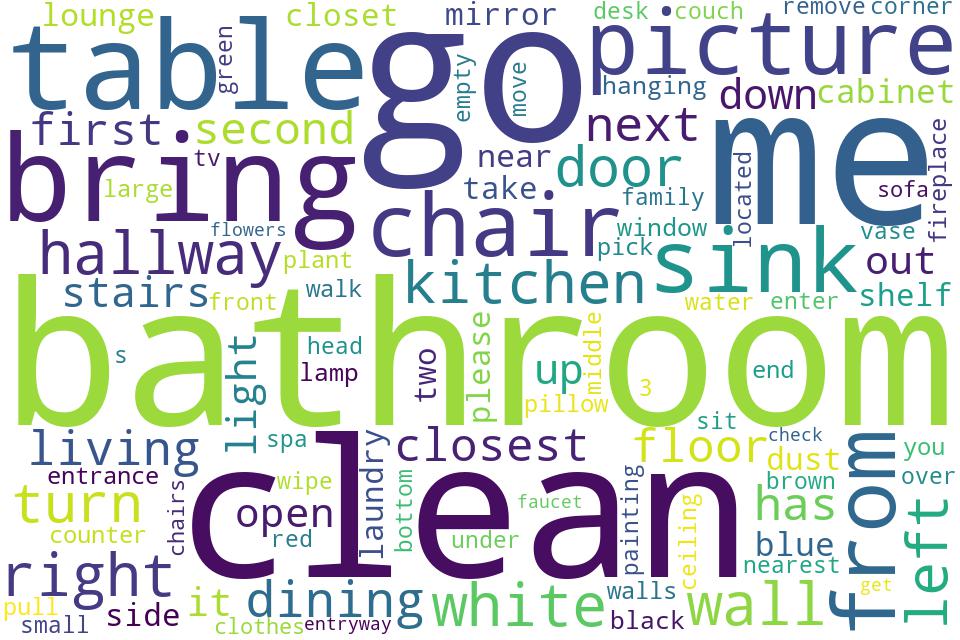}&
         \hspace{-10pt}\includegraphics[width=0.5\linewidth]{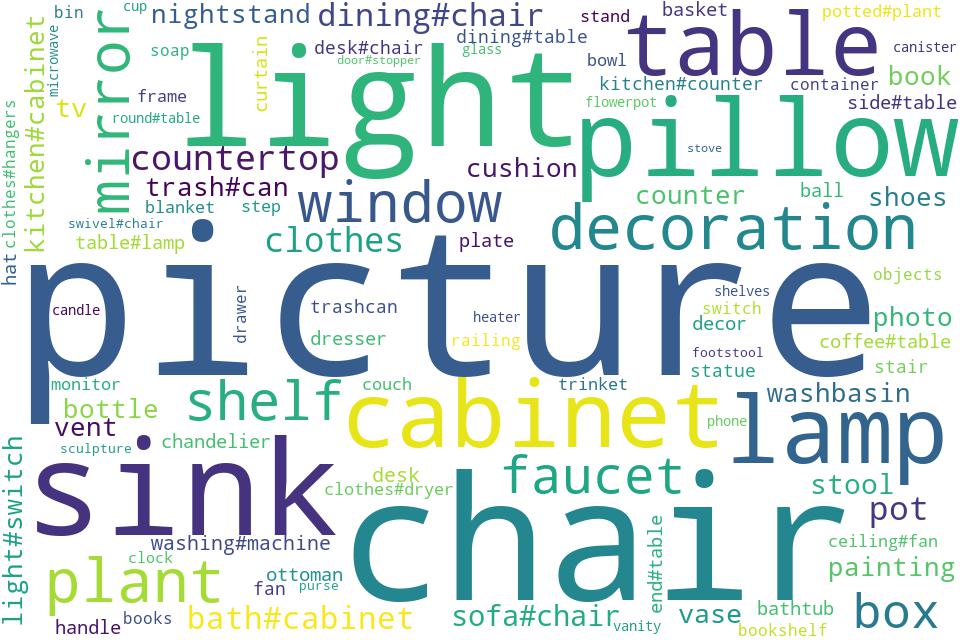}
    \end{tabular}}
\caption{Word cloud of instructions (left) and target objects (right) in the \sexyname{} dataset. The bigger the font, the more percentage it occupies.}
\vspace{-10pt}
\label{fig:wcInstr}
\end{figure}

Fig. \ref{fig:distri} (left) displays the length distribution of the collected instructions, which shows that most instructions have $10\sim 22$ words while the shortest annotation could be only 3 words, such as `Flush the toilet'.
Fig.~\ref{fig:wcInstr} (left) presents the relative amount of words used in instructions in the form of word cloud. It shows that people prefer `go' for navigation, and most instructions involve `bathroom'.
We also compute the number of mentioned objects in instructions and its distribution is presented in Fig.~\ref{fig:distri} (right). 
It shows that $56\%$ instructions mention 3 or more objects, $28\%$ instructions mention 2 objects, and the remaining $15\%$ instructions mention 1 object. On average, there are 7 objects with 50 bounding boxes at each target viewpoint.
There are 4,140 target objects in the dataset, falling into 489 categories, which are 6 times more than the 80 categories in ReferCOCO \cite{YuPYBB16}, a most popular referring expression dataset at present. 
Fig.~\ref{fig:wcInstr} (right)  shows the relative amount of target objects in different categories.

\vspace{-15pt}
\paragraph{Data Splits} We follow the same train/val/test split strategy as the R2R~\cite{r2r} datasets.  The training set consists of 59 scenes and 10,466 instructions over 2,353 objects. The validation set including seen and unseen splits totally contains 63 scenes, 953 objects, and 4,944 instructions, of which 10 scenes and 3,573 instructions over 525 objects are reserved for val unseen split.  For the test set, we collect 6,292 instructions involving 834 objects randomly scattered in 16 scenes. All the test data are unseen during training and validation procedures. The ground truth for the test set will not be released, and we will host an evaluation server where agent trajectories and detected bounding boxes can be uploaded for scoring.

\begin{figure*}[!t]
    \centering
    \hspace{15pt}\includegraphics[width=0.9\linewidth,height=63mm]{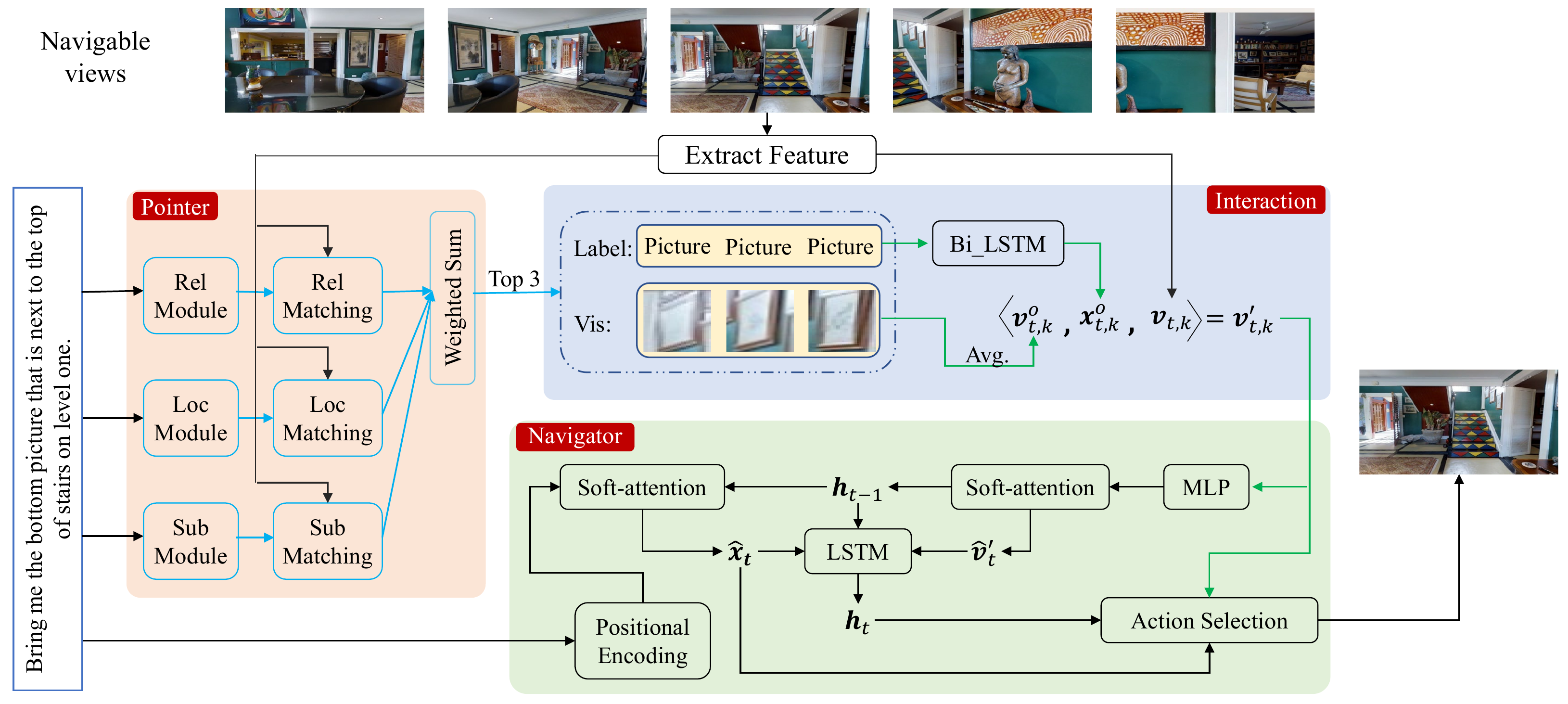}
    \caption{Our Interactive Navigator-Pointer Model}
    \label{fig:my_label}
    \vspace{-8pt}
\end{figure*}

\vspace{-3pt}
\section{The Interactive Navigator-Pointer Model}

\label{sec:ourModel}

As our \sexyname{} task requires an agent to navigate to the goal location and point out the target object, a naive solution is to employ state-of-the-art navigation (as a navigator) and referring expression comprehension (as a pointer) methods jointly.
However, it is of great importance how the navigator and pointer should work together. Ideally, we want the navigator and pointer to benefit each other. Here, we propose a simple yet effective interactive manner that achieves the best performance as a strong baseline.
Specifically, our module includes three parts: a \textit{navigator module} that decides the action to take for the next step, a \textit{pointer module} that attempts to localise the target object according to the language guidance, and an \textit{interaction module} that responses for sending the referring expression comprehension information obtained from the pointer to the navigator to guild it to make more accurate action prediction.

\vspace{-3pt}
\subsection{The Navigator Module}
\vspace{-2pt}
\label{sec:navigator}

The backbone of our navigator module is a `short' version of FAST~\cite{fast}, which uses a sequence-to-sequence LSTM architecture with an attention mechanism and a backtracking mechanism to increase the action accuracy. Specifically, let $\bm{X}\in\mathbb{R}^{L\times512}$ denote instruction features obtained from $\mathcal{X}$ by an LSTM, and $\bm{V'}=[\bm{v'}_{t,1};\dots;\bm{v'}_{t,K}]\in\mathbb{R}^{K\times4736}$ denote updated visual features obtained by our interactive module (Sec.~\ref{sec:interaction}) for panoramic images $\mathcal{V}_t$ at step $t$. FAST-short learns the local logit $l_t$ signal, which is calculated by a visual and textual co-grounding model adopted from~\cite{selfMonitor}.
First, grounded text $\hat{\bm{x}}_t =\bm{\alpha}_t^\top\bm{X}$ and grounded visual $\hat{\bm{v}}'_t=\bm{\beta}_t^\top\bm{V'}$ are learned by
\begin{align}
\bm{\alpha}_t&=\mathrm{softmax}(PE(\bm{X})(\bm{W}_x \bm{h}_{t-1})) \\
\bm{\beta}_t&=\mathrm{softmax}(g(\bm{V'})(\bm{W}_v \bm{h}_{t-1}))
\label{equ:2}
\end{align}
where $\bm{\alpha}_t\in\mathbb{R}^{L\times1}$ is textual attention weight, $\bm{\beta}_t\in\mathbb{R}^{K\times1}$ is visual attention weight, $\bm{W}_x$ and $\bm{W}_v$ are learnable parameters, {$PE(\cdot)$ is the positional encoding~\cite{pe} that captures the relative position between each word within an instruction}, $g(\cdot)$ is a one-layer Multi-Layer Perceptron (MLP), $\bm{h}_{t-1}\in\mathbb{R}^{512\times1}$ is previous encoder context. The new context is updated by an LSTM taking as input the newly grounded text and visual features as well as previous selected action 
\begin{equation}
\label{eq:cellupdate}
    (\bm{h}_t, \bm{c}_t)=\mathrm{LSTM}([\hat{\bm{x}}_t,\hat{\bm{v}}'_t,\bm{a}_{t-1}],(\bm{h}_{t-1}, \bm{c}_{t-1})).
\end{equation}
Then the logit $l_t$ can be computed via an inner-product between each candidate's encoded context and instruction by
\begin{equation}
    l_{t,k} = (\bm{W}_a[\bm{h}_t,\hat{\bm{x}}_t])^\top g(\bm{v'}_{t,k})
    \label{equ:3}
\end{equation}
where $\bm{W}_a$ is a learnable parameter matrix.

Based on logit $l_t$, FAST-short maintains one candidate queue and one ending queue. All navigable viewpoints (including the current viewpoint) at the current location are pushed into the candidate queue, but only the viewpoint with the largest accumulated logit $\sum_{\tau=0}^t{l_\tau}$ is popped out as the selected next step. Each passed viewpoint is pushed into the ending queue. One episode ends if the current viewpoint is selected or the candidate queue is empty or the maximum step is reached. Finally, the viewpoint with the largest accumulated logits is chosen as the actual stop location.

\subsection{The Pointer Module}
\label{sec:pointer}
{
We use MAttNet~\cite{mattnet} as our pointer because of its good generalisation ability. It decomposes an expression into three modular components related to subject appearance, location, and relationship to other objects via the attention mechanism $q^m = \sum_{j=1}^{L}{a_{m,j}e_j},$
where $m\in \{\mathrm{subj, loc, rel}\}$, $e_j$ is the embedding of each word in the expression/instruction $\mathcal{X}$. $a_{m,j}$ is the attention on each word for each module.

Then three kinds of matching scores $S(o_i|q^m)$ are computed for each object $o_i$ conditioned on each modular phrase embedding $q^m$. Specifically, $S(o_i|q^{\mathrm{subj}}) = F(\tilde{v}_i^{\mathrm{subj}},q^{\mathrm{subj}})$, $S(o_i|q^{\mathrm{loc}}) = F(\tilde{l}^{\mathrm{loc}}_i,q^{\mathrm{loc}}),$ and $
S(o_i|q^{\mathrm{rel}})=\max_{j\neq i}F(\tilde{v}^{\mathrm{rel}}_{ij},q^{\mathrm{rel}})$, 
where $F(\cdot)$ is a two-layer MLP, $\tilde{v}_i^{\mathrm{subj}}$ is a `in-box' attended feature for each object using a $14\times 14$ grid. $\tilde{l}_i^{\mathrm{loc}}$ is the location representation of object $o_i$ obtained by a fully-connected layer taking as input the relative position offset and area ratio to its up to five surrounding objects of the same category. $\tilde{v}^{\mathrm{rel}}_{ij}$ is the visual representation of the surrounding object $o_j$ regardless of categories.

The final matching score of object $o_i$ and the instruction $\mathcal{X}$ is a weighted sum:
\vspace{-5pt}
\begin{equation}
\vspace{-5pt}
    S = \sum{S(o_i|q^m)w_m}
\vspace{-3pt}
\end{equation}
where $w_m = \mathrm{softmax}(W_m^L[h_0,h_L]+b_m)$.
}

\subsection{The Interaction Module}
\label{sec:interaction}

Intuitively, we want the Navigator and Pointer to interact with each other so that both navigation and referring expression accuracy can be improved. For example, the navigator can use the visual grounding information to decide when and where to stop, and the pointer accuracy can be improved if the navigator can reach the correct target location. To this end, we propose an interaction module that can plug the pointer's output into the navigator. Specifically, we first perform referring expression comprehension using the above pointer module to select the top-3 matching objects in each candidate view. 
Then we use a trainable bi-direction LSTM to encode the category labels of these selected objects $\mathcal{X}_O=\{\mathrm{Label}_{i\in top3}\}$
\begin{equation}
    \bm{x}_{t,k}^o = \mathrm{bi\_LSTM}(\mathcal{X}_O)
\end{equation}
as the textual representation for the $k$-th candidate viewpoint. In addition, the averaged output of ResNet FC7 layer of these object regions is used as the visual representation $\bm{v}^o_{t,k}$. Finally, we update the candidate viewpoint feature by concatenation 
\begin{equation}
    \bm{v'}_{t,k}= [\bm{v}_{t,k},\bm{x}_{t,k}^o,\bm{v}^o_{t,k}]
\end{equation}
which is send to the navigator (see Equ.~\ref{equ:2} and \ref{equ:3}).
The pointer in such an interaction serves as hard attention for each candidate viewpoint, which highlights the most target-related objects for the navigator to take into account.

\subsection{Loss Functions}
\label{sec:loss}

Our final loss consists of two parts, the navigation loss $L_{nav}$ and referring expression loss $L_{exp}$. The $L_{nav}$ is a cross-entropy loss for action selection and a mean squared error loss for progress monitor:
{\small
\begin{equation}
    L_{nav} = -\lambda_1\sum^{T}_{t=1}{y_{t}^{a}log(l_{t,k})-(1-\lambda_1)\sum^{T}_{t=1}(y_t^{pm}-p_t^{pm})^2}
\end{equation}}
where $y_t^a$ is the ground truth action at step $t$, $\lambda_1=0.5$ is the weight balancing the two loss, $y_t^{pm}\in[0,1]$ is the normalised distance in units of length from the current viewpoint to the goal, $p_t^{pm}$ is the predicted progress.

The referring expression loss $L_{exp}$ is a ranking loss:%
{\small
\begin{equation}
\begin{split}
    L_{exp} = \sum_{i}[\lambda_2\mathrm{max}(0,\delta+S(o_i|r_j)-S(o_i|r_i))\\
    +\lambda_3\mathrm{max}(0,\delta+S(o_k|r_i)-S(o_i|r_i))]
\end{split}
\end{equation}}
where $\lambda_2=1.0$, $\lambda_3=1.0$, $(o_i,r_i)$ is a positive (object, expression) pair, $(o_i,r_j)$ and $(o_k,r_i)$ are negative (object, expression) pairs, $\delta$ is the distance margin between positive and negative pairs. %
All of the losses are summarised together:
 \begin{equation}
     L=L_{nav}+\lambda_4L_{exp}
 \end{equation}
to train our Interactive Navigator-Pointer model. We set $\lambda_4$ to $1.0$ by default.

	\begin{table*}[t]
\centering
\resizebox{\linewidth}{!}{
\begin{tabular}{l|ccccc||ccccc||ccccc}
\hline
\multicolumn{1}{c}{\multirow{3}{*}{Methods}}  & \multicolumn{5}{|c||}{Val Seen} &\multicolumn{5}{c||}{Val UnSeen} & \multicolumn{5}{c}{Test (Unseen)}\\ \cline{2-16}& \multicolumn{4}{c|}{Navigation Acc.}  & REVERIE & \multicolumn{4}{c|}{Navigation Acc.}  & REVERIE   & \multicolumn{4}{c|}{Navigation Acc.}   & REVERIE  \\ \cline{2-5} \cline{7-10} \cline{12-15}  & Succ.& OSucc.   & SPL  & \multicolumn{1}{c|}{Length} & Succ.  & Succ. & OSucc. & SPL &\multicolumn{1}{c|}{Length} & Succ. & Succ.& OSucc. & SPL & \multicolumn{1}{c|}{Length} & Succ.   \\ 
\hline
Random &2.74 &8.92& 1.91 & \multicolumn{1}{c|}{11.99} & 1.97 &1.76& 11.93 &1.01& \multicolumn{1}{c|}{10.76}  & 0.96 &  2.30  &8.88 & 1.44& \multicolumn{1}{c|}{10.34} &1.18 \\
Shortest & 100  & 100 & 100 & \multicolumn{1}{c|}{10.46} &68.45& 100 & 100 & 100 &\multicolumn{1}{c|}{9.47}     & 56.63 & 100 & 100  & 100 & \multicolumn{1}{c|}{9.39} & 48.98    \\
R2R-TF \cite{r2r}&7.38 & 10.75 & 6.40 & \multicolumn{1}{c|}{11.19}  & 4.22 & 3.21& 4.94 &2.80   & \multicolumn{1}{c|}{11.22}  & 2.02 & 3.94 & 6.40 & 3.30  & \multicolumn{1}{c|}{10.07}  & 2.32 \\
R2R-SF \cite{r2r} &29.59&35.70 & 24.01 & \multicolumn{1}{c|}{12.88}  &18.97& 4.20& 8.07 & 2.84   & \multicolumn{1}{c|}{11.07}  & 2.16 & 3.99 & 6.88  & 3.09 & \multicolumn{1}{c|}{10.89}  &2.00\\ \hline
RCM \cite{cross} &23.33& 29.44 & 21.82& \multicolumn{1}{c|}{10.70} & 16.23 & 9.29 & 14.23& 6.97 &\multicolumn{1}{c|}{11.98}     & 4.89 & 7.84  &11.68 & 6.67 & \multicolumn{1}{c|}{10.60} & 3.67\\
SelfMonitor \cite{selfMonitor} & 41.25& 43.29  & 39.61& \multicolumn{1}{c|}{7.54}   & 30.07  & 8.15 & 11.28 &6.44 & \multicolumn{1}{c|}{9.07} & 4.54& 5.80& 8.39 & 4.53 &\multicolumn{1}{c|}{9.23}   & 3.10\\
FAST-Short \cite{fast} & 45.12& 49.68 &40.18& \multicolumn{1}{c|}{13.22}  &31.41 & 10.08 & 20.48 & 6.17  & \multicolumn{1}{c|}{29.70}  & 6.24 & 14.18 & 23.36 & 8.74 & \multicolumn{1}{c|}{30.69}  & 7.07 \\ 
FAST-Lan-Only & 8.36 & 23.61 & 3.67 & \multicolumn{1}{c|}{49.43}  &5.97 & 9.37 & 29.76 & 3.65  & \multicolumn{1}{c|}{45.03}  & 5.00 & 8.15 & 28.45 & 2.88 & \multicolumn{1}{c|}{46.19} & 4.34\\ \hline
\textbf{Ours} & \textbf{50.53} & \textbf{55.17}  & \textbf{45.50} & \multicolumn{1}{c|}{{16.35}}  & \textbf{31.97} & \textbf{14.40} & \textbf{28.20}  & \textbf{7.19} & \multicolumn{1}{c|}{{45.28}}  &\textbf{7.84} & \textbf{19.88} & \textbf{30.63} & \textbf{11.61} & \multicolumn{1}{c|}{{39.05}}  & \textbf{11.28} \\ \hline
Human & -- & --  & -- & \multicolumn{1}{c|}{--}& -- & -- & --  & --  & \multicolumn{1}{c|}{--}     & -- & 81.51 & 86.83  & 53.66 & \multicolumn{1}{c|}{21.18} & 77.84\\ \hline
\end{tabular}}
\vspace{0mm}
\caption{REVERIE success rate achieved  by combining state-of-the-art navigation methods with the RefExp method MAttNet~\cite{mattnet}.}
\label{tab:finaltab}
\vspace{-3mm}
\end{table*}

\section{Experiments}
\label{exp}

In this section, we first present the training details of the interactive navigator--pointer model. Then, we provide extensive evaluation and analysis.

\subsection{Implementation Details}
 
The simulator image resolution is set to 640$\times$480 pixels with a vertical field of view of 60 degrees. 
For each instruction in the train split,  images and object bounding boxes at the goal viewpoint (for the views where the target object is visible)  are organised following the format as in MAttNet for pointer training. With the trained pointer, assistant object information is provided for the navigator as described in Section~\ref{sec:interaction} to train the navigator. The code and dataset will be released.

\subsection{\sexyname Experimental Results}

We first evaluate several baseline models and state-of-the-art (SoTA) navigation models, combined with the MattNet, \ie, the pointer module.
{After the navigation models decide to stop, the pointer module is used to predict the target object. In addition, we also test human performance using an interactive web interface (see details in the supplementary).

Below is a brief introduction of the evaluated baseline and state-of-the-art models. }There are four baseline models, which are:
\begin{itemize}
\vspace{-8pt}
\item{\textbf{Random}} exploits the characteristics of the dataset by randomly choosing a path with random steps (maximum 10) and then randomly choose an object as the predicted target.
\vspace{-8pt}
\item{\textbf{Shortest}} always follows the shortest path to the goal. %
\vspace{-8pt}
\item{\textbf{R2R-TF and R2R-SF ~\cite{r2r}}} are the first batch of navigation baselines, which trains a basic LSTM~\cite{lstm} with attention mechanism~\cite{attnbengio}. The difference between R2R-TF and R2R-SF is that R2R-TF is trained with the ground truth action at each step (Teacher-Forcing, TF) while R2R-SF adopts an action sampled from the predicted probability over its action space (Student-Forcing, SF). %
\vspace{-5pt}
\end{itemize}

The evaluated four SoTA navigation models are:
\begin{itemize}
\vspace{-8pt}
\item{\textbf{SelfMonitor~\cite{selfMonitor}}} uses a visual-textual co-grounding module to highlight the instruction for the next action and a progress monitor to reflect the progress.
\vspace{-8pt}
\item{\textbf{RCM~\cite{cross}}} employs reinforcement learning to encourage global matching between instructions and trajectories, and performs cross-model grounding.
\vspace{-8pt}
\item{\textbf{FAST-Short~\cite{fast}}} introduces backtracking into SelfMonitor.
\vspace{-8pt}
\item{\textbf{FAST-Lan-Only}} employs above FAST-Short model but we only input the language instruction without any visual input. {This model is used to check whether our task/dataset has a bias on language input.}
\vspace{-5pt}
\end{itemize}

\noindent{\textbf{Results.}}
The detailed experimental results are presented in Tab.~\ref{tab:finaltab}, of which the first four rows are results for baselines, the following four rows are for SoTA methods, and the last two rows are for our model and human performance.

According to the baseline section in Tab.~\ref{tab:finaltab}, the Random model only achieves a \sexyname{} success {around} $1\%$, which indicates the \sexyname{} task has a huge solution space. The sequence-to-sequence baselines R2R-TF and R2R-SF \cite{r2r} achieve good results on the Val-Seen split but decrease  a lot on the unseen splits. Student-Forcing is generally better than Teacher-Forcing. The Shortest model achieves the perfect performance because the ground-truth path to the goal is directly given.

In the second part, the best \sexyname{} success rate is achieved by the combination of SoTA navigation (FAST) and referring expression (MAttNet) models. However, the  \sexyname{} success rate is only $7.07\%$ on the test split, falling far behind human performance $77.84\%$. The navigation-only accuracy of these SoTA navigation models indicates the challenge of our navigation task. Nearly $30\%$ drops on the unseen splits are observed compared to the performance on previous R2R \cite{r2r}. For example, the navigation SPL score of FAST-Short~\cite{fast} on Val-UnSeen split drops from $43\%$ on the R2R dataset to $6.17\%$ on \sexyname. 

To test whether our dataset has strong language bias, \ie, whether a language-only model can achieve good performance, we implement a FAST-Lan-Only model with only instructions as its input. We observe a big drop on both seen and unseen splits, which suggests jointly considering language and visual information is necessary to our task.

Overall, these SoTA model results show that a simple combination of SoTA navigation  and referring expression methods would not necessarily lead to the best performance as failures from either the navigator or the pointer would decrease the overall success. 
In this paper, we make the first attempt to enable the navigator and pointer to work interactively as described in Sec.~\ref{sec:interaction}. The results in Tab.~\ref{tab:finaltab} show that our interactive model achieves consistently better results than non-interactive ones. The FAST-Short can be treated as our ablated model that without our proposed interaction module. Our final model achieves a gain of $4.2\%$ on the test split. 

\vspace{-8pt}
\paragraph{Referring Expression-Only.} We also report the Referring Expression-Only performance. In this setting, the agent is placed at the ground-truth target location, and then referring expression comprehension models are tested. 

We test the SoTA models such as MattNet~\cite{mattnet} and CM-Erase~\cite{cm-erase} as well as a simple CNN-RNN baseline model with triplet ranking loss. Tab.~\ref{tab:grounding-only} presents the results with human performance.
It shows that the SoTA models achieve around $50\%$ accuracy on the test split\footnote{These SoTA models achieve $80\%$ accuracy on ReferCOCO \cite{YuPYBB16}, a golden benchmark for referring expression.} which are far more better than the results when jointly considering the navigation and referring expression shown in Tab. \ref{tab:finaltab}. Even though, there is still a $40\%$ gap to human performance, suggesting that our proposed \sexyname{} task is challenging.

\begin{table}[t]
	\centering
	\small
	\vspace{1mm}
	\begin{tabular}{lccc}
		\hline
		& Val Seen & Val UnSeen& Test \\
		\hline
		Baseline &30.69&18.63& 16.18 \\
		MAttNet \cite{mattnet} & 68.45&56.63 & 48.98 \\
		CM-Erase \cite{cm-erase} &65.21&54.02 & 45.25 \\
		Human & -- & -- & 90.76\\
		\hline
	\end{tabular}%
	\vspace{1mm}
	\caption{Referring expression comprehension success rate ($\%$) at the ground truth goal viewpoint of our \sexyname{} dataset.}
	\vspace{-3mm}
	\label{tab:grounding-only}%
\end{table}%

	\section{Conclusion}

Enable human-robots collaboration is a long-term goal.
In this paper, we make a step further towards this goal by proposing a Remote Embodied Visual referring Expressions in Real Indoor Environments (\sexyname{}) task and dataset.  The \sexyname{} is the first one to evaluate the capability of an agent to follow high-level natural languages instructions to navigate and identify the target object in previously unseen real images rendered buildings. We investigate several baselines and an interactive Navigator-Pointer agent model, of which the performance consistently demonstrate the significant necessity of further researches in this field.
		
We reach three main conclusions: First, \sexyname{} is interesting because existing vision and language methods can be easily plugged in. Second, the challenge of understanding and executing high-level instructions is significant. Finally, the combination of instruction navigation and referring expression comprehension is a challenging task due to the large gap to human performance.
	
	\section{Acknowledgments}
	We thank Sam Bahrami and Phil Roberts for their great help in the building of the REVERIE dataset.

	{\small
		\bibliographystyle{ieee_fullname}
		\bibliography{egbib}
	}
	\clearpage
	
	\setcounter{section}{0}
	\setcounter{figure}{0}
	\setcounter{table}{0}
	\setcounter{footnote}{0}
	\setcounter{equation}{0}
	{\noindent\large\textbf{Supplements}}
\vspace{3mm}

In this supplementary material, we provide detailed explanation of evaluation metrics, examples of collected data, data collecting tools, how human test are performed and visualisation of several REVERIE results.

\section{Evaluation Metrics}
\begin{itemize}
	\item Navigation Success: a navigation is considered successful only if the target object can be observed at the stop viewpoint. Please note that to encourage the agent to approach closer  to the target object, we set the objects visible if they are within 3 meters away from the current location. 
	\item Navigation Oracle Success: a navigation is considered oracle successful if the target object can be observed at one of its passed viewpoints.
	\item Navigation SPL: it is the navigation success weighted by the length of navigation path, which is 
		\begin{equation}
			\frac{1}{N}\sum_{i=1}^{N}{S_i\frac{\ell_i}{\max(\ell_i,p_i)}}
		\end{equation}
	where $N$ is the number of tasks, $S_i\in\{0,1\}$ is a binary indicator of success of task $i$, $\ell_i$ is the shortest length between the starting viewpoint and the goal viewpoint of task $i$, and $p_i$ is the path length of an agent for task $i$.
	\item Navigation Length: trajectory length in meters.
	\item REVERIE Success: a task is considered REVERIE successful if the output bounding box has an IoU (intersection over union) $\geq0.5$ with the ground truth. 
\end{itemize}
\section{Typical Samples of The \sexyname{} Task}
\begin{figure*}[!htbp]
	\centering
	\includegraphics[width=0.8\linewidth]{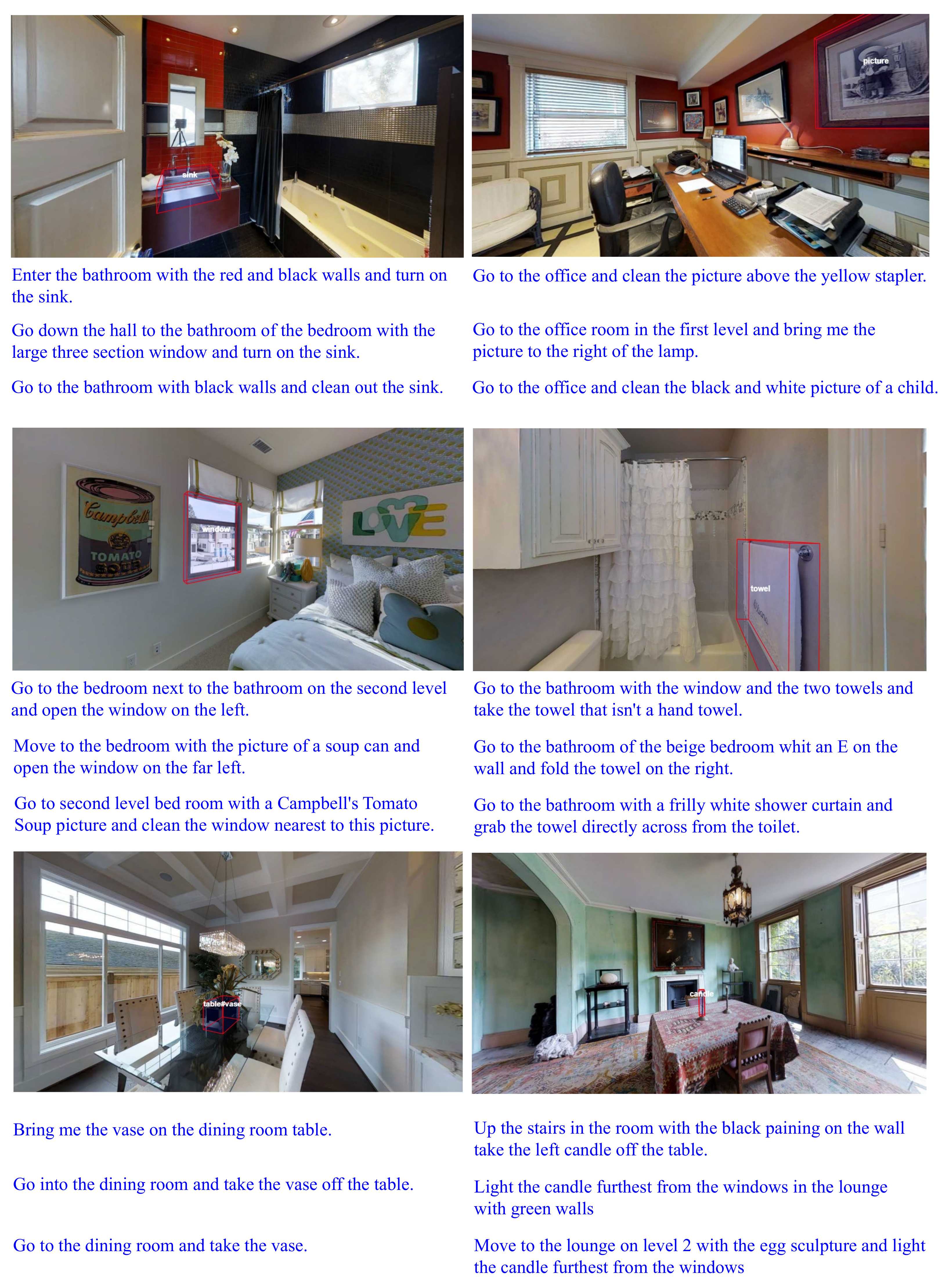}
	\caption{Several typical samples of the collected dataset, which involves various object category, goal region, path instruction, and object referring expression.}
	\label{fig:samples}
\end{figure*}

 In Figure~\ref{fig:samples}, we present several typical samples of the proposed \sexyname{} task. It shows the diversity in object category, goal region, path instruction, and target object referring expression.

\section{Data Collecting Tools}
\begin{figure*}[!b]
	\centering
	\includegraphics[width=\linewidth]{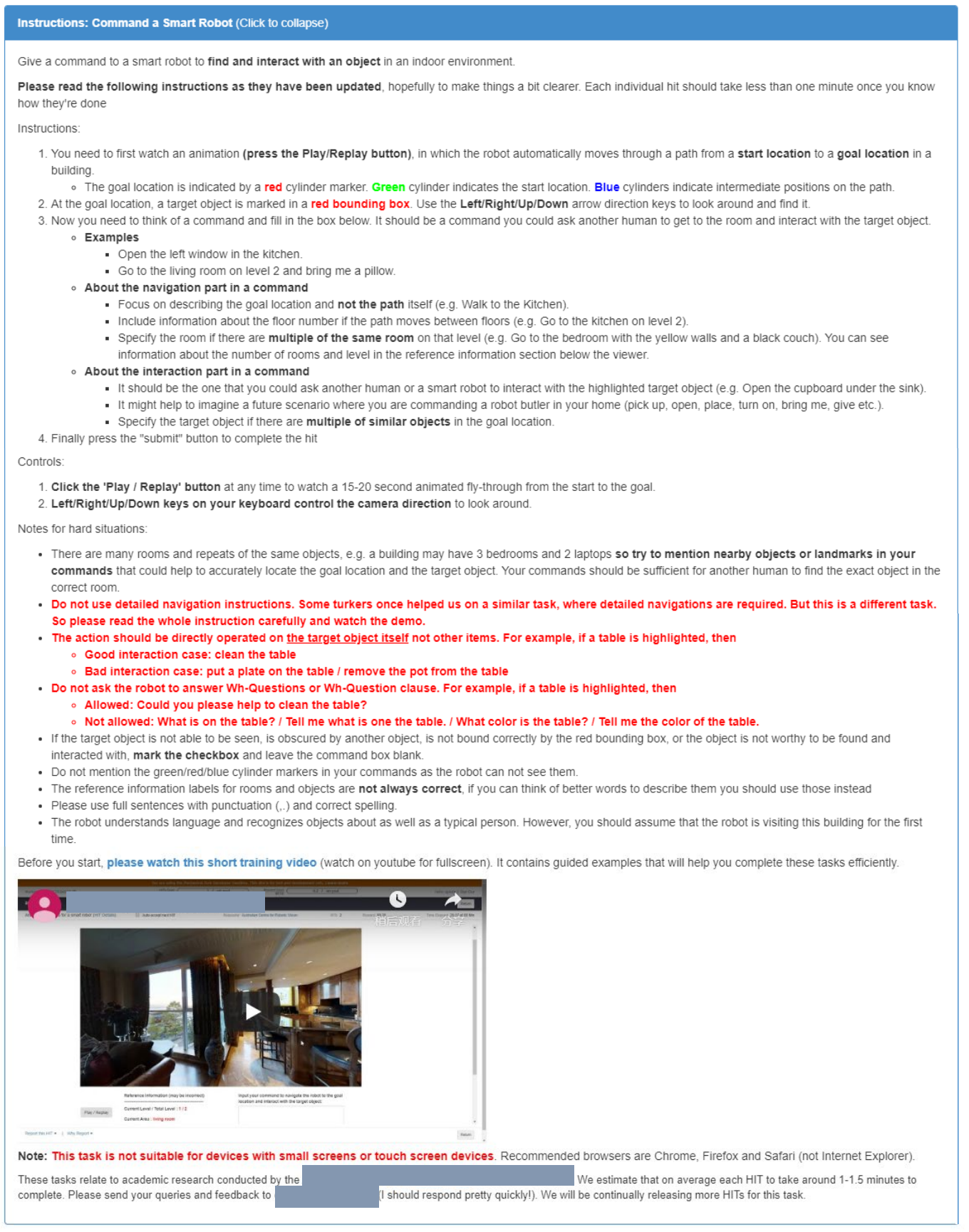}\\
	\caption{Data collecting interface part I: instructions for AMT workers. }
	\label{fig:instruction}
\end{figure*}

\begin{figure*}[t]
	\centering
	\includegraphics[width=\linewidth]{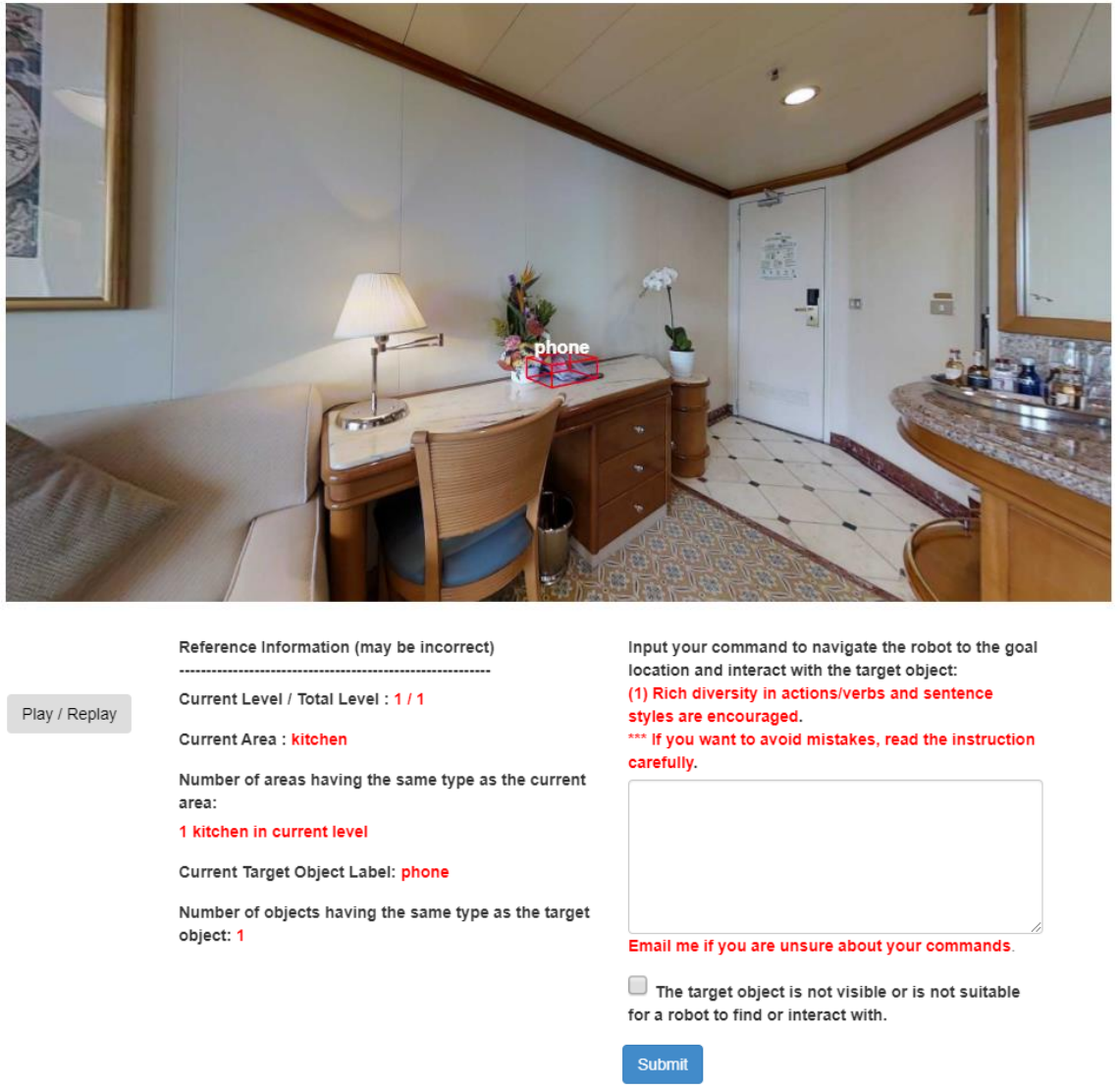}\\
	\caption{Data collecting interface Part II: assistant information and user input field.}
	\label{fig:submit}
\end{figure*}

To collect data for the \sexyname{} task, we develop a WebGL based data collecting tool as shown in Figure~\ref{fig:instruction} and Figure~\ref{fig:submit}. To facilitate the workers, we provide real-time updated reference information in the web page according to the location of the agent, including the current level/total level, the current region, and the number of regions in the build having the same function. At the goal location, in addition to highlighting the target object with a red 3D rectangle, we also provide the label of the target object and the number of objects falling in the same category with the target object.  Text and video instructions are provided for workers to understand how to make high quality annotations as shown in Figure~\ref{fig:instruction}. 

\section{Human Performance Test}
\begin{figure*}[t]
	\centering
	\includegraphics[width=\linewidth]{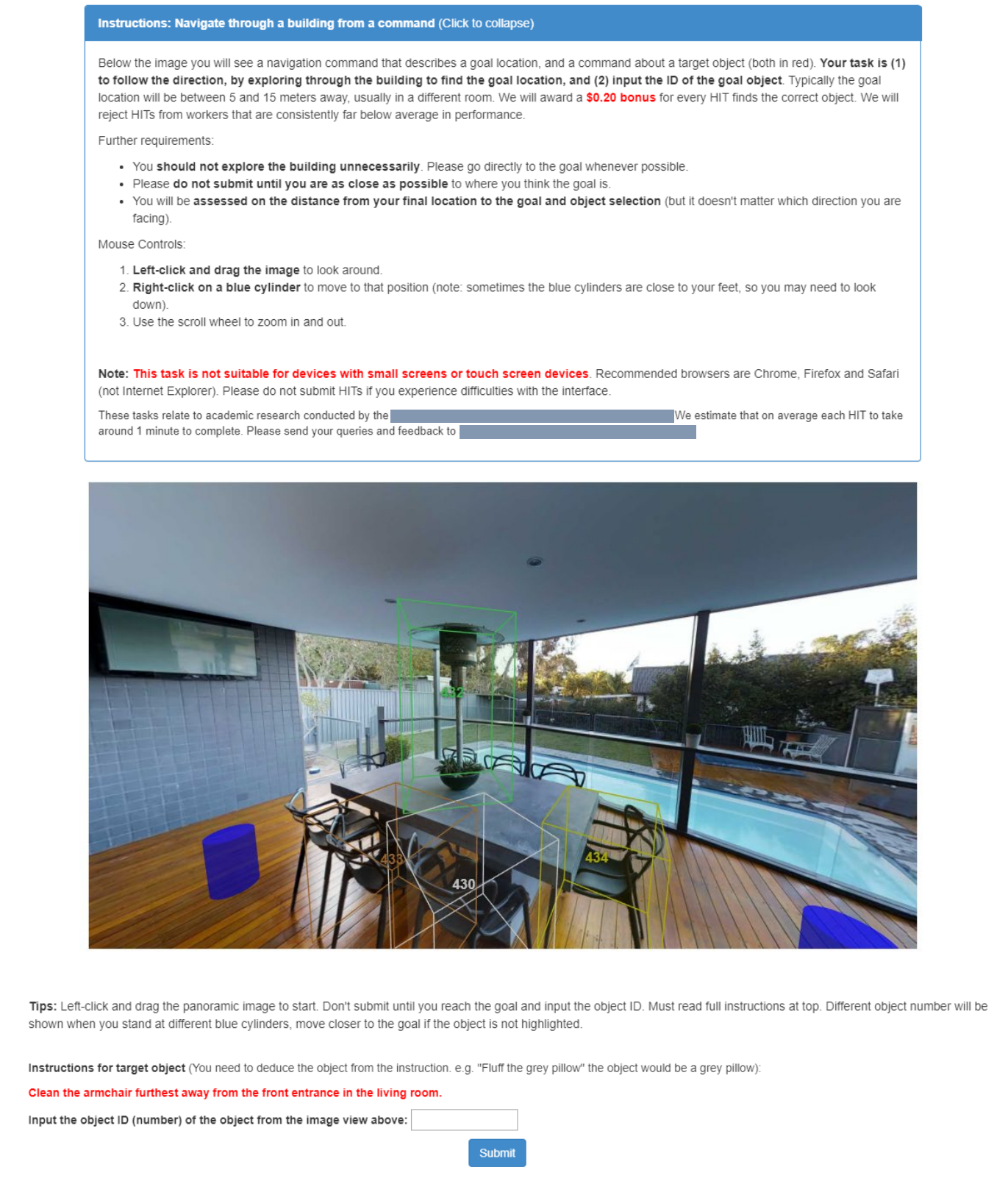}\\
	\caption{Human test interface. Workers need first to navigate to the goal location by clicking or dragging mouse, and then identify the target object. Only objects within 3 meters from the current location are highlighted. Different colors are used to facilitate workers to distinguish one object from others.}
	\label{fig:eval}
\end{figure*}

To obtain the machine-human performance gap, we develop a WebGL based tool as shown in Figure~\ref{fig:eval} to test human performance. In the tool, we show an instruction about a remote object to the worker. Then the worker needs to navigate to the goal location and select one object as the target object from a range of object candidates. The worker can look around and go forward/backword by dragging or clicking.

\section{Visualisation of REVERIE Results}
\begin{figure*}[!htbp]
	\centering
	\includegraphics[width=0.83\linewidth]{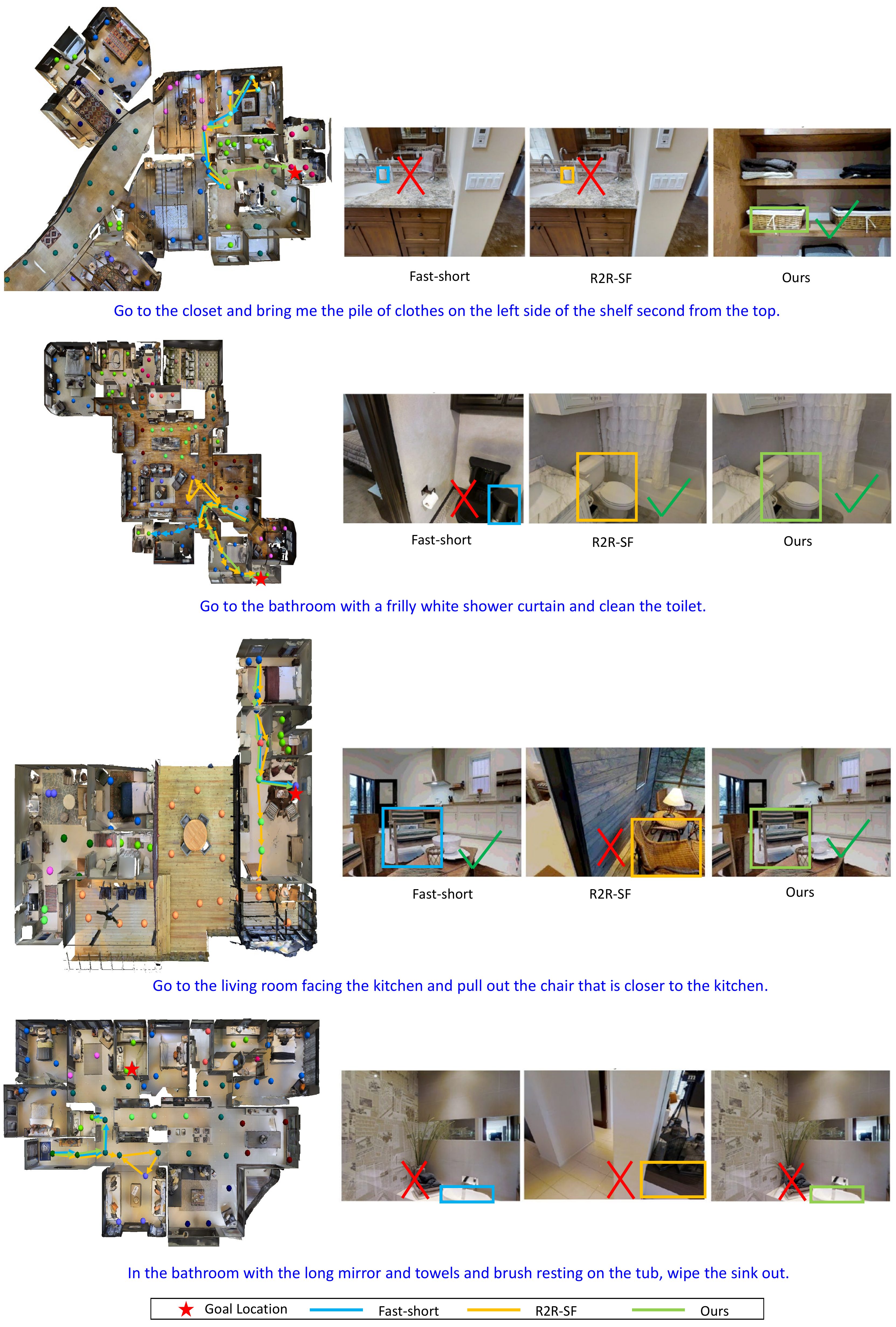}
	\caption{Visualisation of several REVERIE results, including trajectories and referring expression grounding of three typical methods. Colorized dots denote reachable locations, and different colors mark locations belonging to different regions according to the Matterport dataset.}
	\label{fig:vis}
\end{figure*}

In Figure~\ref{fig:vis}, we provide the visualisation of several REVERIE results obtained by the typical state-of-the-art method, FAST-short, and the typical baseline method, R2R-SF.

 
	\end{document}